\documentclass{article}
\usepackage{spconf,amsmath,graphicx}

\usepackage{lipsum}
\usepackage{placeins}
\usepackage[font=footnotesize,skip=0.5em]{caption} 

\title{A Prototypical Triplet Loss for Cover Detection}
\twoauthors
  {Guillaume Doras}
	{Sacem \& Ircam Lab,\\
	 CNRS, Sorbonne Universit\'e\\
	guillaume.doras@sacem.fr}
  {Geoffroy Peeters}
	{LTCI, T\'el\'ecom Paris\\
	Institut Polytechnique de Paris\\
	geoffroy.peeters@telecom-paris.fr}

\begin{document}
\ninept
\maketitle

\begin{abstract}
Automatic cover detection -- the task of finding in an audio dataset all covers of a query track -- has long been a challenging theoretical problem in MIR community. It also became a practical need for music composers societies requiring to detect automatically if an audio excerpt embeds musical content belonging to their catalog.

In a recent work, we addressed this problem with a convolutional neural network mapping each track's dominant melody to an embedding vector, and trained to minimize cover pairs distance in the embeddings space, while maximizing it for non-covers. We showed in particular that training this model with enough works having five or more covers yields state-of-the-art results.

This however does not reflect the realistic use case, where music catalogs typically contain works with zero or at most one or two covers. We thus introduce here a new test set incorporating these constraints, and propose two contributions to improve our model's accuracy under these stricter conditions: we replace dominant melody with multi-pitch representation as input data, and describe a novel prototypical triplet loss designed to improve covers clustering. We show that these changes improve results significantly for two concrete use cases, large dataset lookup and live songs identification.

\end{abstract}
\vspace{0.5em}
\section{Introduction}
\label{sec:intro}

Covers are different interpretations of the same original musical work. They usually share a similar melodic line, but typically differ greatly in one or several other dimensions, such as structure, tempo, key, instrumentation, genre, etc. Automatic cover detection -- the task of finding in an audio corpus all covers of one or several query tracks -- has long been seen as a challenging theoretical problem in MIR. It also became an acute practical problem for music composers societies facing continuous expansion of user-generated online content including musical excerpts under copyright.

Cover detection is not \textit{stricto sensu} a classification problem: due to the ever growing amount of musical works (the classes) and the relatively small number of covers per work (the samples), the actual question is not so much ``to which work this track belongs \mbox{to ?}'' as ``to which other tracks this track is the most similar ?''. Formally, cover detection therefore requires to establish a similarity relationship between a query track and a reference track: it implies the composite of a feature extraction function preserving common musical facets between different covers of the same work, followed by a simple pairwise comparison function allowing fast lookup in large music corpora. 

In a recent work, we proposed such a solution based on a convolutional neural network designed to map each track's dominant melody representation to a single embedding vector. We trained this network to minimize the Euclidean distance between cover pairs embeddings, while maximizing it for non-cover pairs \cite{doras2019cover}. We showed that such a model trained with works having enough covers can learn a similarity metric between dominant melodies. We showed in particular that training this model with works having at least five to ten covers yields state-of-the-art accuracy on small and large scale datasets.

This does not, however, reflects the realistic use case: in practice, the reference dataset is usually a music catalog containing one original interpretation of each work, along with typically zero or more rarely one or two covers. Moreover, query tracks have no particular reason to be covers of any work seen during training. 

We thus extend here our earlier work as follows: we first build a new test set reflecting realistic conditions, i.e. disjoint from the original training set and containing only works having very few covers. Our main contribution to improve our model's generalization under these stricter conditions is then twofold: we propose the use of a multi-pitch representation instead of dominant melody as input data, and we introduce a novel prototypical triplet loss designed to improve covers clustering. We finally show that these changes improve results significantly on two common tasks in the industry: large dataset lookup and live songs identification.

In the rest of this paper, we review in \S\ref{sec:related_work} the main concepts used in this work. We describe our proposed improvements, namely the multi-pitch representation and the prototypical triplet loss in \S\ref{sec:method}. We detail our large dataset lookup experiment and its results in \S\ref{sec:experiments_1}, and the live song identification experiment and its results in \S\ref{sec:experiments_2}. We conclude with future improvements to bring to our method.

\section{Related work}
\label{sec:related_work}

\subsection{Multi-pitch estimation}
\label{subsec:multi_pitch_extraction}
Dominant melody and multi-pitch estimation have long been another challenging problem in the MIR community \cite{klapuri2006multiple, salamon2012melody}. A major breakthrough was brought recently with the introduction of a convolutional network that learns to extract the dominant melody out of the audio Harmonic CQT \cite{bittner2017deep}. 

We built upon this idea in a recent work and proposed for this task an adaptation of U-Net -- a model originally designed for medical image segmentation \cite{ronneberger2015u, doras2019use} -- which yields state-of-the-art results.

\begin{figure}[h!]
\centerline{
\includegraphics[width=0.49\columnwidth]{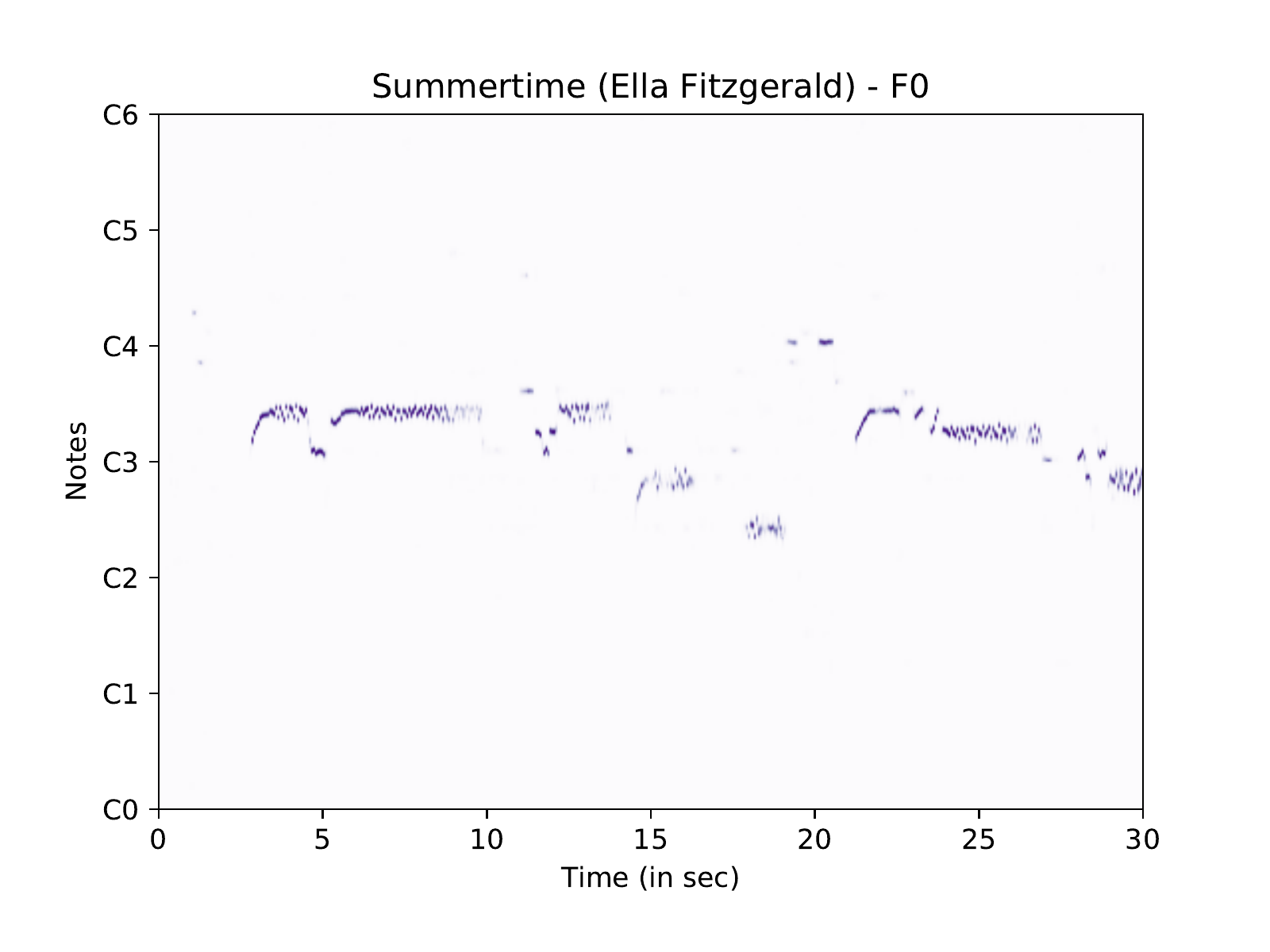}
\includegraphics[width=0.49\columnwidth]{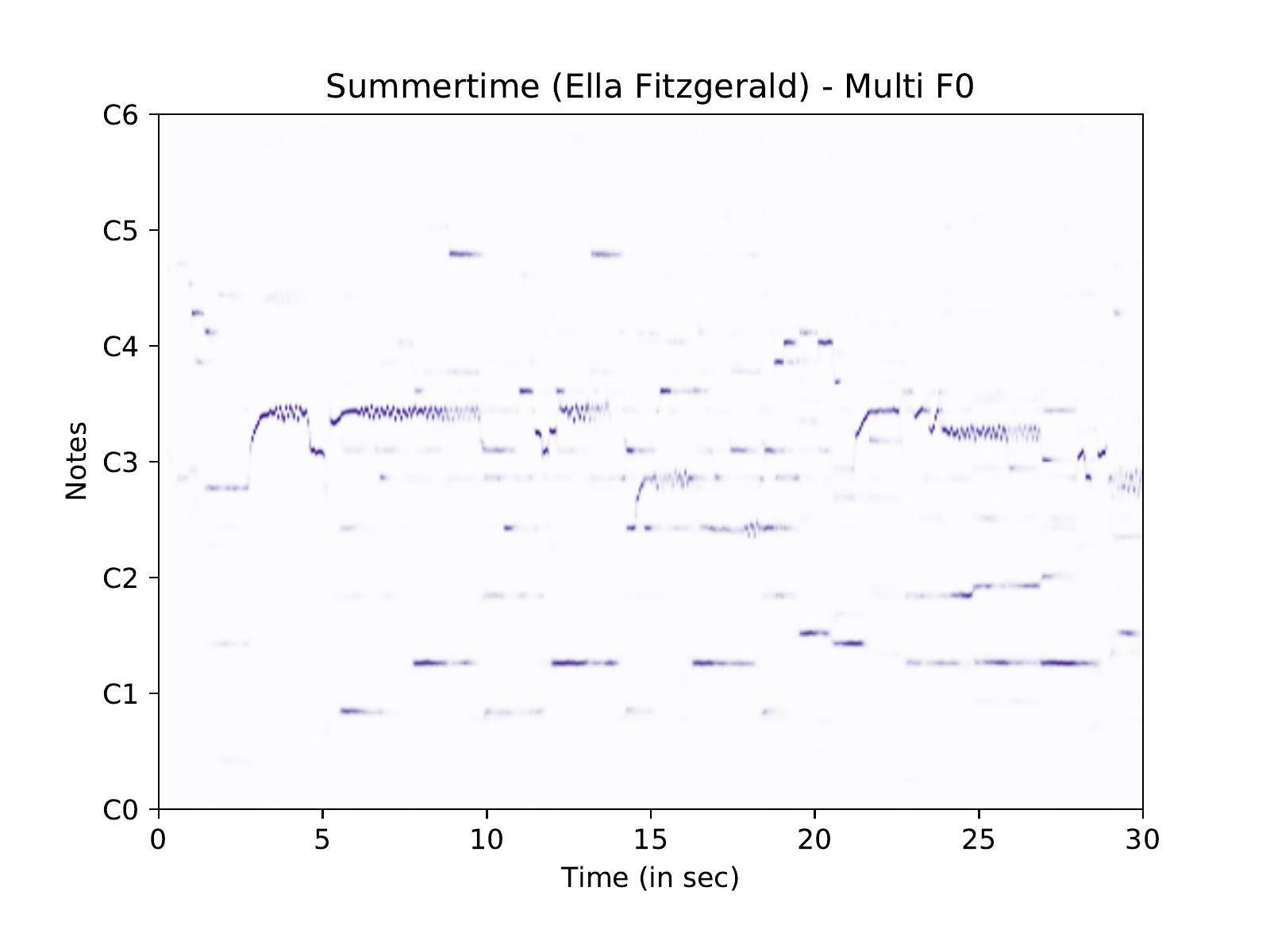}
}
 \caption{Dominant melody (left) and multi-pitch (right) estimated by our U-Net \cite{doras2019use} for the first 30 sec. of Ella Fitzgerad's interpretation of Summertime.}
 \label{fig:summertime_ella_fitzgerald}
\end{figure}

A multi-pitch representation typically contains the dominant melody, and usually includes also the bass line and some extra isolated melodic lines, as can be seen on Fig. \ref{fig:summertime_ella_fitzgerald} which shows the two representations of the same track obtained by two versions of our U-Net trained for each task.

In the following, we propose to compare cover detection results obtained when using dominant melody vs. multi-pitch as input data. Our assumption behind this proposal is that multi-pitch embeds useful information shared by covers that is not present in the dominant melody.

\vspace{-1.5em}

\subsection{Cover detection}
\label{subsec:related_work_cover_detection}

Successful approaches in cover detection usually first extract an input representation preserving common musical facets between different versions -- in particular dominant melody \cite{sailer2006finding, marolt2008mid, tsai2008using} or harmonic/tonal  progression \cite{ellis2007identifyingcover, serra2009cross, bello2007audio, salamon2013tonal, silva2016simple}, and then compute a similarity score between pairs of melodic and/or harmonic sequences, typically using a cross-correlation \cite{ellis2007identifyingcover, ravuri2010cover} or an alignment scoring algorithm \cite{gomez2006song, serra2009cross}.

To alleviate the cost of the comparison function, various authors proposed to encode each audio representation as a single scalar or vector -- its embedding, and to reduce the similarity computation to a simple Euclidean distance between embedding. This approach shifts the computation burden to the audio features extraction function -- which can be done offline and stored, and allows faster lookup of large dataset. Originally, embeddings were hashing functions of the original representation \cite{bertin2012large, marolt2008mid}, but as for many other MIR applications, ad-hoc -- and somewhat arbitrary -- hand-crafted features extraction was progressively replaced with data-driven automatic metric learning \cite{humphrey2012moving, humphrey2013data, tsai2016known, qi2018triplet}. 

In the following, we extend in particular our earlier work describing a model learning a similarity metric between dominant melody representations \cite{doras2019cover}.

\subsection{Similarity metric learning}
\label{subsec:similarity_metric_learning}

Learning a similarity metric that generalizes to unseen samples is a common task in machine learning \cite{bellet2013survey}. Particularly relevant to our problem, the seminal Nearest Component Analysis \cite{goldberger2005neighbourhood} learns a Mahalanobis distance metric that is used to improve k-nearest neighbors classification. It directly inspired Nearest Class Mean algorithm which performs classification based on the distance to the mean of the class instead of each sample in the class \cite{mensink2013distance}.
Similarly, Large Margin Nearest Neighbor \cite{weinberger2006distance} learns a metric based on triplet samples to enforce each sample of a class to be closer to other members of its class than to any sample of another class.

These ideas were generalized to learn not only the parametric part of a Mahalanobis distance but instead an end-to-end transformation of an input data to its representation in an embedding space. In particular, the \textit{center} loss was proposed in \cite{wen2016discriminative} to minimize intra-class variations and to maximize different classes separation, while triplet mining was proposed to improve the original triplet loss \cite{schroff2015facenet}.

Similar approaches have also been proposed in the related context of few-shot learning, where the model shall generalize to samples of potentially unknown class \cite{lake2011one, koch2015siamese}. In particular, prototypical networks compute for each sample query a class probability distribution based on its distance with the centroid or \textit{prototype} of each class, and is trained to maximize the probability of the correct class \cite{snell2017prototypical}. 

In the following, we build upon these previous works and introduce the \textit{prototypical} triplet loss, which represents a class by its prototype rather than by the set of its samples.

\section{Proposed method}
\label{sec:method}

We now extend our earlier work replacing dominant melody with multi-pitch as input data, and adapting the standard triplet loss to a novel training loss designed to improve covers clustering.

\subsection{Realistic test set}
\label{sec:method_test_set}

We built for our previous work a dataset SHS$_{\text{5+}}$ of covers provided by the SecondHandSong website API \footnote{http://secondhandsongs.com}. This initial dataset contains about 7.5k works with a least five covers per work, for a total of 62k tracks, i.e. 8.2 covers per work on average\footnote{\label{note:covers_dataset}The dataset is available at https://gdoras.github.io/topics/coversdataset}. This high amount was required for efficient training of our model, but does not mimic a realistic music catalog. 

We thus built for this paper another set SHS$_{\text{4--}}$, totally disjoint from the original training set, containing only the works provided by SecondHandSong with exactly two, three or four covers per work. This new set contains about 49k songs for about 20k works, i.e. 2.5 covers per work on average. It reflects better a realistic music catalog, and implies harder testing conditions: having less covers per work implies more non-cover confusing tracks for each query. In the following, we have thus used SHS$_{\text{4--}}$ as our new test set\footnotemark[2].

\subsection{Multi-pitch as input data}
\label{sec:input_data}

\begin{figure}[h!]
  \centering
  \centerline{\includegraphics[width=0.9\columnwidth]{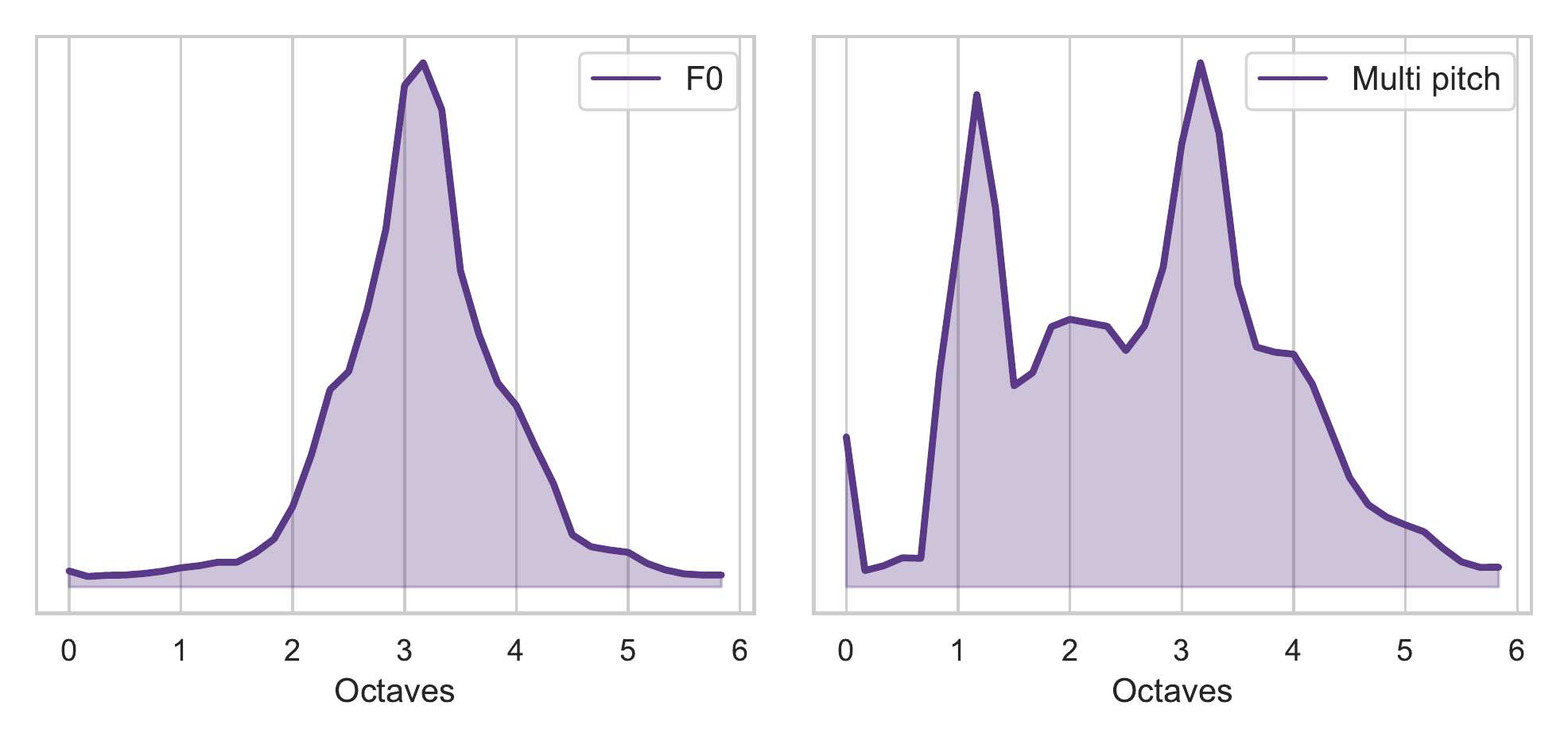}}
\caption{Dominant melody (left) and multi-pitches (right) tone distributions for the 62k tracks of the original SHS$_{\text{5+}}$ set.}
\label{fig:f0_vs_multi_pitches_distributions}
\end{figure}

Dominant melody and multi-pitch representations are extracted from audio HCQT across 6 octaves with a frequency resolution of 5 bins per semi-tone , as in \cite{bittner2017deep,doras2019use}. A dominant melody typically does not exhibit important variations and its frequency range can be trimmed to 3 octaves around its mean pitch value with no loss of information, as seen in \cite{doras2019cover}. Conversely, multi-pitch values are spread across several octaves with two major modes corresponding to the bass lines and the dominant melody, as can be seen on Fig. \ref{fig:f0_vs_multi_pitches_distributions}.

In this work, we thus trimmed each multi-pitch representation to 5 octaves centered around its mean pitch value. As done in \cite{doras2019cover}, only the first 180 seconds are considered, and the resulting matrix is scaled down by a factor 5 to one bin per semi-tone resolution for a final shape of $1024 \times 60$ bins.

\subsection{Prototypical triplet loss}
\label{subsec:prototypical_triplet_loss}

We denote by $C=\{C_i\}_{i\in 1..|C|}$ the set of classes, and denote by $S_{C_i}=\{s_{j_i}\}_{j_i \in 1..|S_{C_i}|}$ the set of samples of class $C_i$ -- here, classes are musical works, and samples are covers of a work.

The triplet loss is used to train a model to map each sample to an embedding closer to all of its positive counterparts than it is to all of its negative counterparts. Formally, for all triplets \{$a$, $p$, $n$\} where $a$ is an anchor, and $p$ or $n$ is one of its positive or negative example, respectively, the loss to minimize is expressed as $\ell=\max(0, d_{\mathrm{ap}} + \alpha - d_{\mathrm{an}})$, where $\alpha$ is a margin and $d_{\mathrm{ap}}$ and $d_{\mathrm{an}}$ are the distances between each anchor $a$ and $p$ or $n$, respectively \cite{weinberger2006distance,schroff2015facenet}.

\begin{figure}[!h]
\centerline{
\includegraphics[width=0.9\columnwidth]{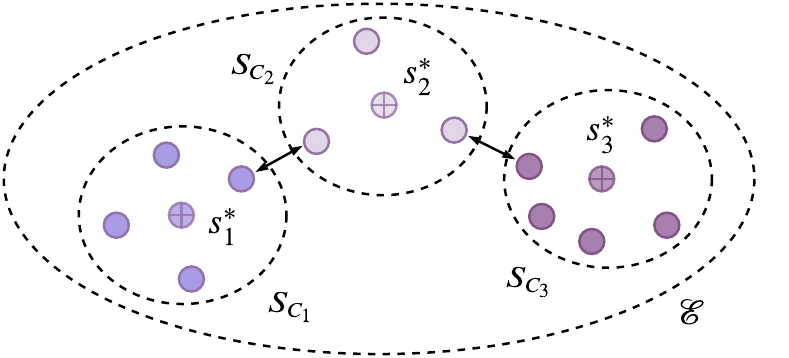}
}
 \caption{Edging points with a black arrow are closer to each other than they are to some points of their own cluster, but are closer to the centroid of their cluster than to centroids of other classes.}
 \label{fig:embeddings_centroids}
\end{figure}

In our previous work, we used a triplet loss with semi-hard negatives mining to cluster together covers of the same work. It appeared that tracks incorrectly paired as covers on the test set were generally located at the edge of their cluster, and could be closer to samples belonging to other classes than to some samples of their own class, as illustrated on Fig. \ref{fig:embeddings_centroids}. 

In order to improve samples clustering around their class centroid, \cite{wen2016discriminative, snell2017prototypical} proposed in a different context to represent a class by the centroid of its samples rather than by the set of its samples. We propose to follow this idea in the triplet loss context, and introduce here the \textit{prototypical} triplet loss.

Consider an anchor sample $s^a_{c} \in S_{C}$, and the triplet \{$s^a_{c}$, $s^{*}_{c}$, $s^{*}_{c'\neq c}$\}, where $s^{*}_{c}$ is the centroid of the cluster $S_{C}$ of $s^a_{c}$, and where $s^{*}_{c'\neq c}$ is the centroid of another cluster $S_{C'}$. We denote $d_{\mathrm{ap}^{*}}$, resp. $d_{\mathrm{an}^{*}}$, the Euclidean distance between $s^a_{c}$ and $s^{*}_{c}$, resp. $s^{*}_{c'\neq c}$, and define the prototypical triplet loss as $\ell=\max(0, d_{\mathrm{ap}^{*}} + \alpha - d_{\mathrm{an}^{*}})$.

In practice, prototype embeddings are computed online at each training step: let $(\mathbf{e}_j)_{j\in 1..B}$ be the embeddings matrix of samples belonging to classes $C=\{C_i\}_{i \in 1..|C|}$ present in each batch of size $B$. The prototype embeddings matrix $(\mathbf{e^*}_i)$ is then simply:
\begin{equation*}
    (\mathbf{e^*}_i) = (\delta_{ij}) (\mathbf{e}_j) \quad i,j \in 1..|C| \times 1..B 
\end{equation*}

\noindent where $\delta_{ij}=1/|S_{C_i}|$ if $s_j\in S_{C_i}$ and 0 otherwise\footnote{See implementation at https://github.com/gdoras/PrototypicalTripletLoss}.

Semi-hard triplets mining is then conducted online for each anchor in the batch among the available \{$s^a_{c}$, $s^{*}_{c}$, $s^{*}_{c'\neq c}$\} triplets, as in \cite{schroff2015facenet}. The positions of each sample in a class and the centroid of this class are learned jointly to minimize their distance in the embedding space, while maximizing it to all other cluster centroids. 

Once trained, the centroid of the embeddings of the covers of a work can be considered to be the embedding of the work itself, which interestingly embodies the concept of musical work.

\section{Large dataset lookup experiment}
\label{sec:experiments_1}

This experiment corresponds to the use case where the work and/or covers of a query track shall be found in a reference music corpus. 

\subsection{Description}
\label{subsec:experiments_1_description}

We split SHS$_{\text{4--}}$ set into a query set picking randomly one cover per work (i.e. 20k queries), and kept entire SHS$_{\text{4--}}$ as reference set. We extracted all embeddings with our model trained on SHS$_{\text{5+}}$. We computed the 20k$\times$49k pairwise distance matrix between queries and references, discarding self pairs.

We report here standard information retrieval measures, used in particular in MIREX cover song identification task\footnote{https://www.music-ir.org/mirex/wiki/2019}: the Mean Average Precision (MAP) gives an insight on how all correct samples are ranked, while the mean number of true positives in the top ten positions (MT@10) indicates the relevance of the ranking. 

\begin{figure}[h!]
\centerline{
\includegraphics[width=0.49\columnwidth]{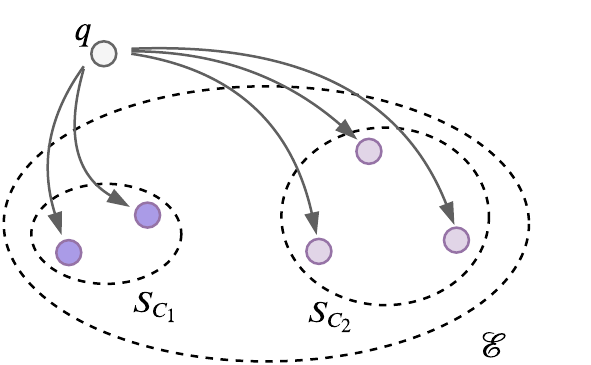}
\includegraphics[width=0.49\columnwidth]{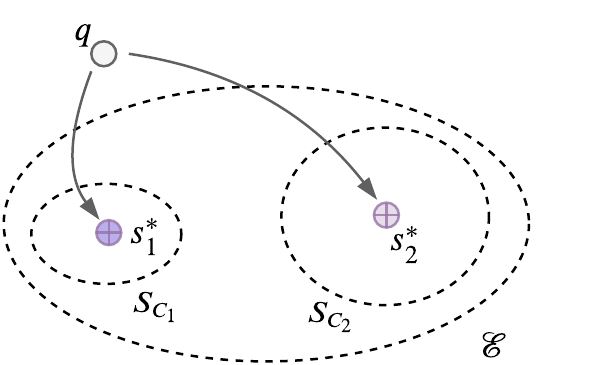}
}
 \caption{Computing distances between query and samples (left) or between query and class prototypes (right) corresponds to a scoring scheme by samples or by classes.}
 \label{fig:scoring_schemes}
\end{figure}

We considered here two scoring schemes illustrated on Fig. \ref{fig:scoring_schemes}: \textit{by samples} compares each query with each reference as usual, while \textit{by class} compares each query with the \textit{prototype} of each class. Scoring by samples makes sense when looking for \textit{covers} of a track, while scoring by classes makes sense when looking for the \textit{work} of a track.
\vspace{-0.3em}

We expect that results will be mechanically better with the latter scheme, as there are less class prototypes than samples to compare with. For a fair comparison, we also reported the normalized MT@$10^*$, i.e MT@10 divided for each query by the maximum number of correct samples that it can return.

\subsection{Results}
\label{subsec:experiments_1_results}

\begin{figure}[h!]
  \centering
  \centerline{\includegraphics[width=0.95\columnwidth]{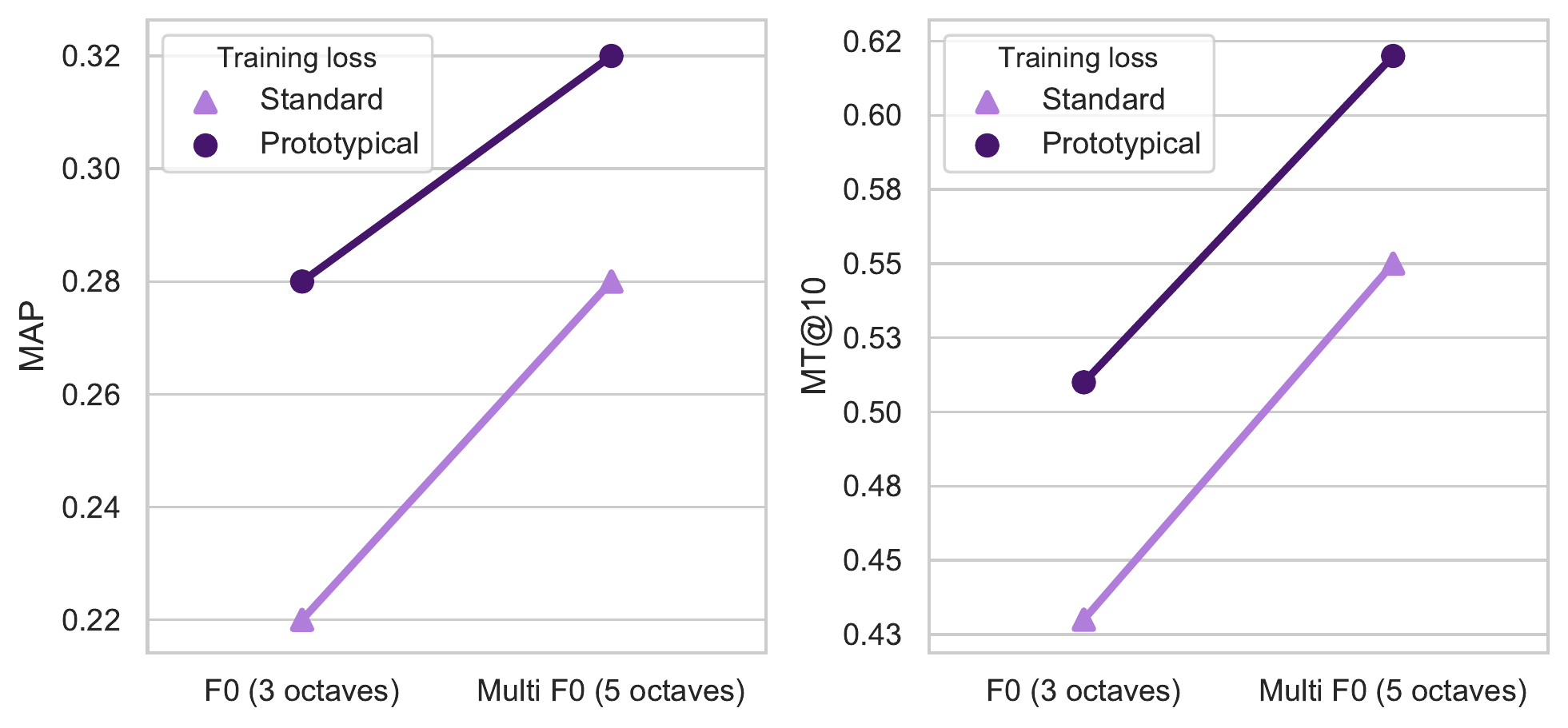}}
\caption{MAP (left) and MT@10 (right) on 20k queries vs. 49k references for a model feed with dominant melody vs. multi-pitch and trained with standard (pink) vs. prototypical (purple) triplet loss.}
\label{fig:scores_comparison_shs_2-3-4_1-all_scored_samples}
\end{figure}

Averaged scores by samples obtained on ten random splits with dominant melody vs. multi-pitch representations are depicted on Fig. \ref{fig:scores_comparison_shs_2-3-4_1-all_scored_samples}.

\FloatBarrier
It appears both for the standard and prototypical triplet loss that multi-pitch representation improves all scores compared to dominant melody. This suggests that dominant melody alone does not embed all relevant information for cover detection, while multi-pitch extra information (mainly the bass line) helps to improve results further.

\begin{figure}[h!]
\centerline{\includegraphics[width=1.0\columnwidth]{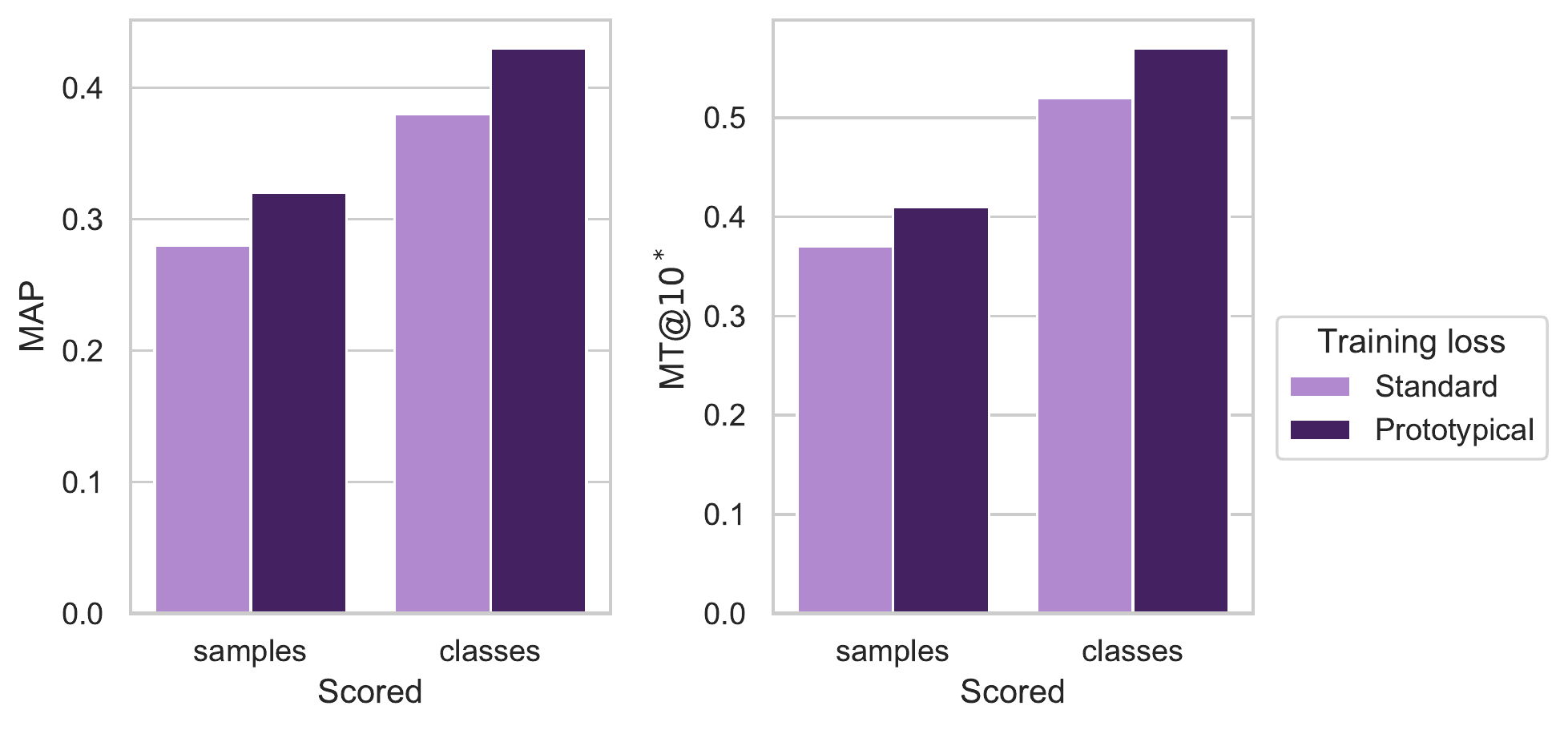}}
\caption{MAP (left) and normalized MT@10$^*$ (right) obtained for 20k queries vs. 49k references for a model fed with multi-pitch, trained with standard (pink) vs. prototypical (purple) triplet loss, and scored in samples and classes modes.}
\label{fig:scores_comparison_bars_shs_2-3-4_1-all_f0}
\end{figure}

Averaged scores for multi-pitch for both schemes obtained on ten random splits with standard vs. prototypical triplet loss are depicted on Fig. \ref{fig:scores_comparison_bars_shs_2-3-4_1-all_f0}.

\FloatBarrier

It appears that all scores are significantly improved by 5 to 6\% when the model is trained with the prototypical triplet loss. This was expected when scoring in \textit{classes} mode, as the model has been trained specifically for this task. But even more interestingly, results are also significantly improved when the model is trained with a prototypical triplet loss and used in \textit{samples} mode to look for all covers of a given query. 

In other words, training the model with the prototypical triplet loss improves generalization to unseen examples for both lookup schemes, by samples or by classes.

\section{Live song identification experiment}
\label{sec:experiments_2}

This experiment corresponds to the use case where songs played at a live concert shall be found in a reference music catalog. 

\subsection{Description}
\label{subsec:experiments_2_description}

In this experiment, we used five concert recordings belonging to a proprietary collection of the french music composers society (Sacem). Each recording lasts between one to three hours and contains 10 to 30 songs. Audio quality is usually poor, as recording is done from the audience. 

We split each concert recording into 180s. overlapping frames with a 30s. hop, and extract the embedding of each frame with our trained model. Concert frames embeddings are used as the query set.

Similarly, we extracted the embedding of a studio version for each of the $N$ songs played during each concert, and added them to the 49k embeddings of SHS$_{\text{4--}}$. We computed the distance matrix between queries (concert overlapping frames) vs. the references (original version tracks + 49k confusing tracks). At each time frame, only the best reference is considered, and only references showing up for at least three consecutive time frames are kept. 

We report here the R-precision obtained on the candidates list, i.e. the number of correct tracks found at rank $N$ divided by $N$.

\subsection{Results}
\label{subsec:experiments_2_results}

\begin{figure}[h!]
\centerline{\includegraphics[width=1.0\columnwidth]{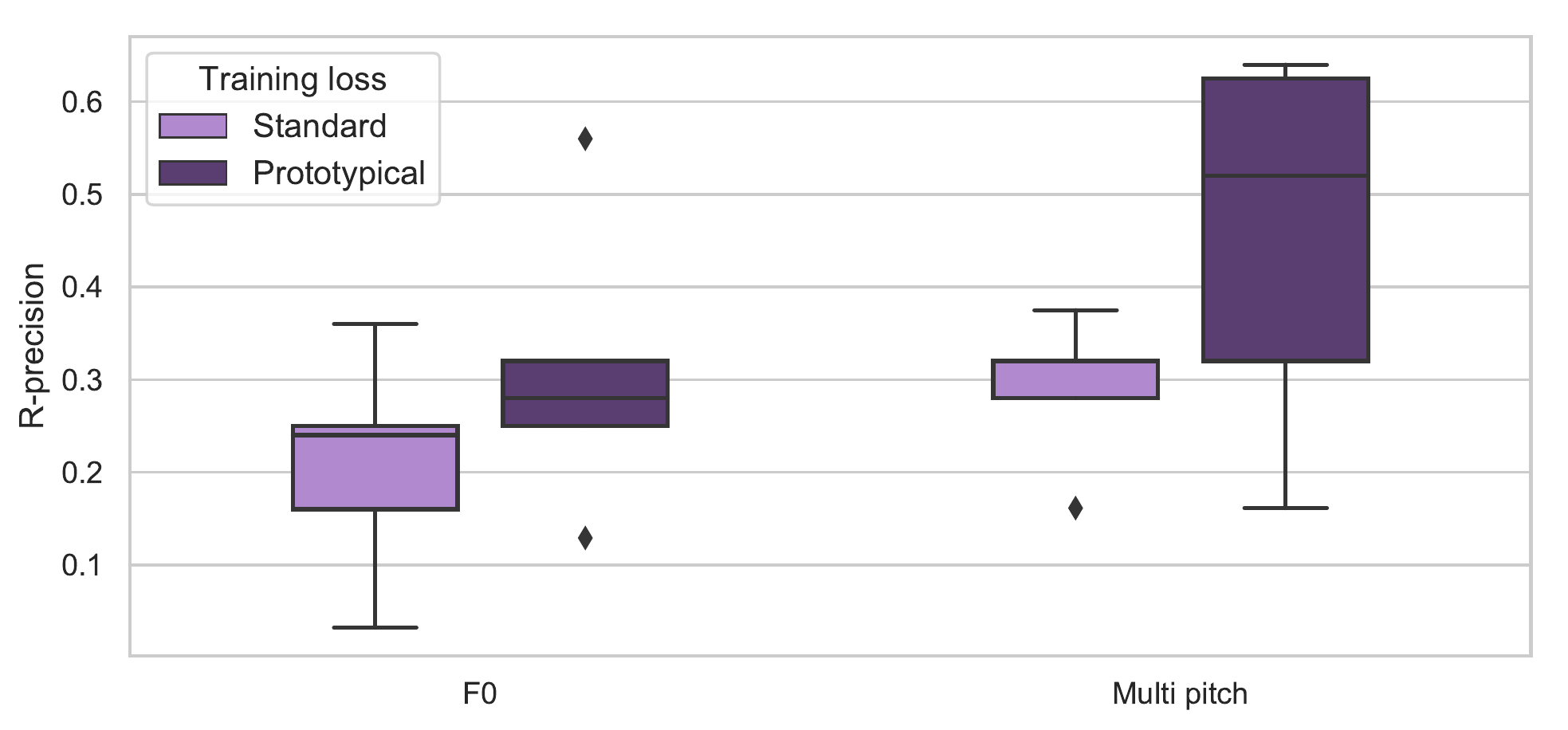}}
\caption{R-precision on five concerts for a model feed with dominant melody (left) and multi-pitch (right), trained with standard (pink) vs. prototypical (purple) triplet loss.}
\label{fig:r_precisions}
\end{figure}

The R-precision obtained for live songs identification with different models are shown on Fig. \ref{fig:r_precisions}. As for previous experiment, it clearly shows that multi-pitch representation improves results compared to dominant melody, probably because bass lines are more salient in live recording conditions.

It confirms that the prototypical triplet loss improves scores compared to standard triplet loss. All in all, the model trained with prototypical triplet loss on multi-pitch input data yields the best performance.
\vspace{-0.5em}
\vspace{-0.3em}

\begin{figure}[h!]
\centerline{\includegraphics[width=1.0\columnwidth]{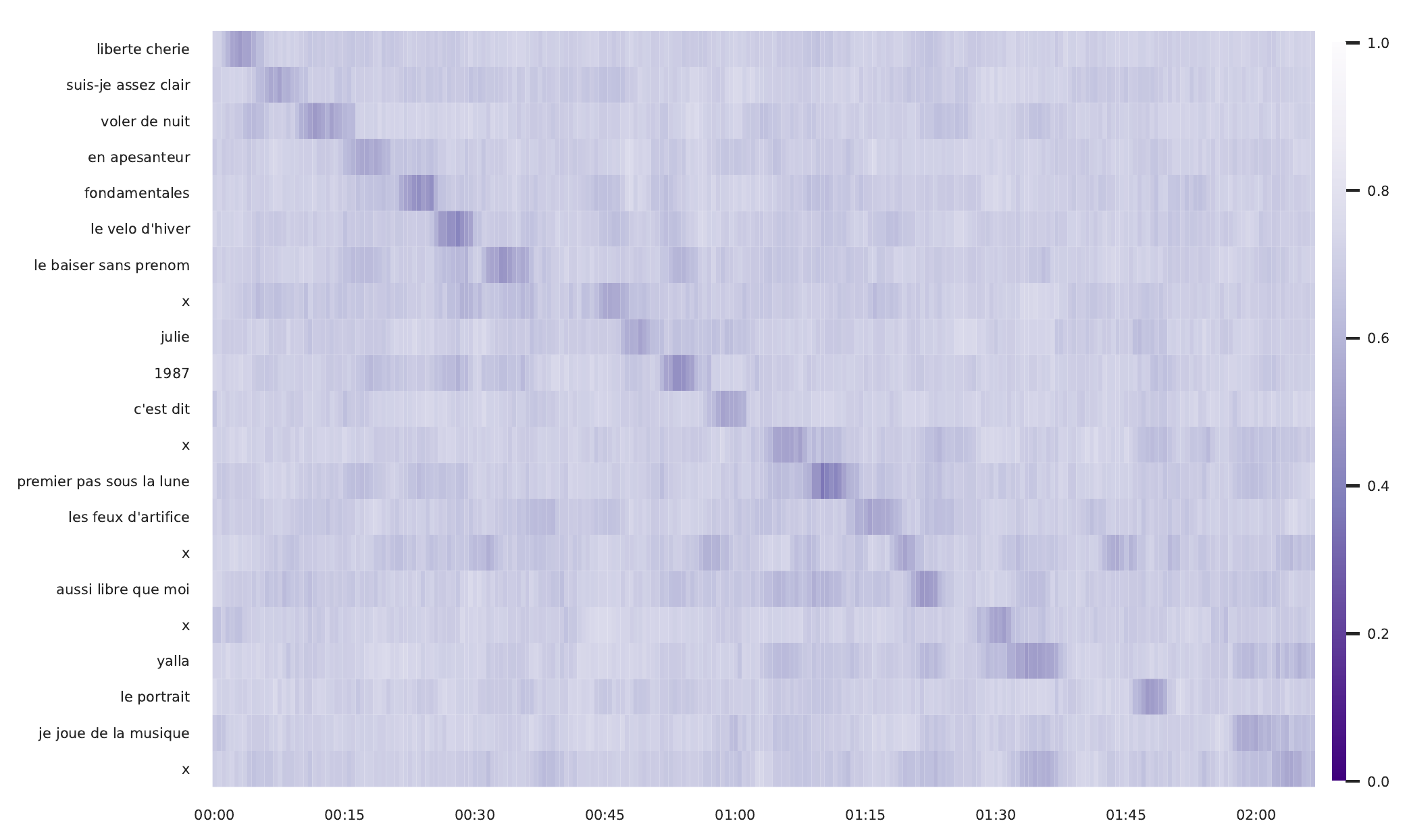}}
\caption{Best matching tracks at each frame for a live concert of French artist Calogero (darker is closer). Original titles on Y-axis (wrongly returned tracks are marked with 'x'), time in hours:minutes on X-axis. R-precision=0.64}
\label{fig:concert}
\end{figure}

As an illustration, the distances matrix between best candidate tracks embedding and each time frame embedding is shown on Fig.\ref{fig:concert} for the concert that yield to the best R-precision.
\vspace{-0.7em}

\section{Conclusion}
\label{sec:conclusion}

We proposed two improvements to our previous state of the art cover detection model: the use of multi-pitch representation instead of dominant melody as input data, and the \textit{prototypical} triplet loss, an adaptation of the standard triplet loss designed to improve the clustering of samples around the centroid or prototype of their class. 

We introduced a new 50k tracks test set reflecting realistic usage conditions in the industry, and showed that these two modifications significantly improve the results compared to our previous setup for two tasks -- large dataset lookup and live songs identification.

We however acknowledge that our test set is still smaller than the ones used in the industry by several order of magnitude, and our future work will address this question.


\vfill\pagebreak


\bibliographystyle{IEEEbib}
\bibliography{refs}

\end{document}